\documentclass[conference]{IEEEtran}
\IEEEoverridecommandlockouts

\usepackage{listings} 
\usepackage{changepage} 
\usepackage[figurename=Fig., tablename=Tab., small]{caption}
\usepackage{subcaption}
\usepackage{algorithm2e}[ruled]
\SetKwComment{Comment}{/* }{ */}
\RestyleAlgo{ruled}

\usepackage{cite}
\usepackage{amsmath,amssymb,amsfonts}
\usepackage{algorithmic}
\usepackage{graphicx}
\usepackage{textcomp}
\usepackage{xcolor}

\usepackage[utf8]{inputenc}
\usepackage{pgfplots}
\DeclareUnicodeCharacter{2212}{−}
\usepgfplotslibrary{groupplots,dateplot}
\usetikzlibrary{patterns,shapes.arrows}
\pgfplotsset{compat=newest}

\def\BibTeX{{\rm B\kern-.05em{\sc i\kern-.025em b}\kern-.08em
    T\kern-.1667em\lower.7ex\hbox{E}\kern-.125emX}}

\usepackage{pgfplots}
\usetikzlibrary{babel}
\usetikzlibrary{calc}
\usepgfplotslibrary{groupplots}

\pgfplotsset{compat=1.16,
    tick label style = {font = {\fontsize{6pt}{12pt}\selectfont}},
    label style = {font = {\fontsize{8pt}{12pt}\selectfont}},
    legend style = {font = {\fontsize{8pt}{12pt}\selectfont}},
    title style = {font = {\fontsize{8pt}{12pt}\selectfont}},
  }
\usepackage{pgfplotstable}

\newcommand{\linebreakand}{%
  \end{@IEEEauthorhalign}
  \hfill\mbox{}\par
  \mbox{}\hfill\begin{@IEEEauthorhalign}
}

\begin{document}

\title{Eye Sclera for Fair Face Image Quality Assessment}

\author{Wassim Kabbani \quad Kiran Raja \quad  Raghavendra Ramachandra
\quad  Christoph Busch\\
Norwegian University of Science and Technology, Gjøvik, Norway \\
\small \{wassim.h.kabbani; kiran.raja; raghavendra.ramachandra; christoph.busch \} @ntnu.no
}

\IEEEoverridecommandlockouts \IEEEpubid{\makebox[\columnwidth]{979-8-3503-5447-8/24/\$31.00 ©2024 IEEE \hfill} \hspace{\columnsep}\makebox[\columnwidth]{ }}

\maketitle

\IEEEpubidadjcol

\begin{abstract}
Fair operational systems are crucial in gaining and maintaining society's trust in face recognition systems (FRS). FRS start with capturing an image and assessing its quality before using it further for enrollment or verification. Fair Face Image Quality Assessment (FIQA) schemes therefore become equally important in the context of fair FRS. This work examines the sclera as a quality assessment region for obtaining a fair FIQA. The sclera region is agnostic to demographic variations and skin colour for assessing the quality of a face image. We analyze three skin tone related ISO/IEC face image quality assessment measures and assess the sclera region as an alternative area for assessing FIQ. Our analysis of the face dataset of individuals from  different demographic groups representing different skin tones indicates sclera as an alternative to measure dynamic range, over- and under-exposure of face using sclera region alone. The sclera region being agnostic to skin tone, i.e., demographic factors, provides equal utility as a fair FIQA as shown by our Error-vs-Discard Characteristic (EDC) curve analysis. 

\end{abstract}

\begin{IEEEkeywords}
FIQA, Face Recognition, Fair Facial Biometrics, Eye Sclera, Skin-tone, Exposure, Dynamic Range
\end{IEEEkeywords}

\section{Introduction}

The quality of face images presented to a face recognition system (FRS) heavily influences its performance \cite{b30}. Thus, it is important to control face image quality before enrolment of reference samples and before the verification attempt. Face Image Quality Assessment (FIQA) is a process that measures the quality of a face image in terms of its utility for face recognition. The NIST FRVT report \cite{b9} states - \textit{"With good quality portrait photos, the most accurate algorithms will find matching entries, when present, in galleries containing 12 million individuals, with rank one miss rates of approaching 0.1\%"}. A face image quality score can be an overall unified quality score that does not necessarily depend on specific features of the image explicitly or individual quality components that assess a particular aspect of the face image independently from others.

\begin{figure}[h!]
     \centering
     \begin{subfigure}[b]{0.35\linewidth}
         \centering
         \includegraphics[width=\linewidth]{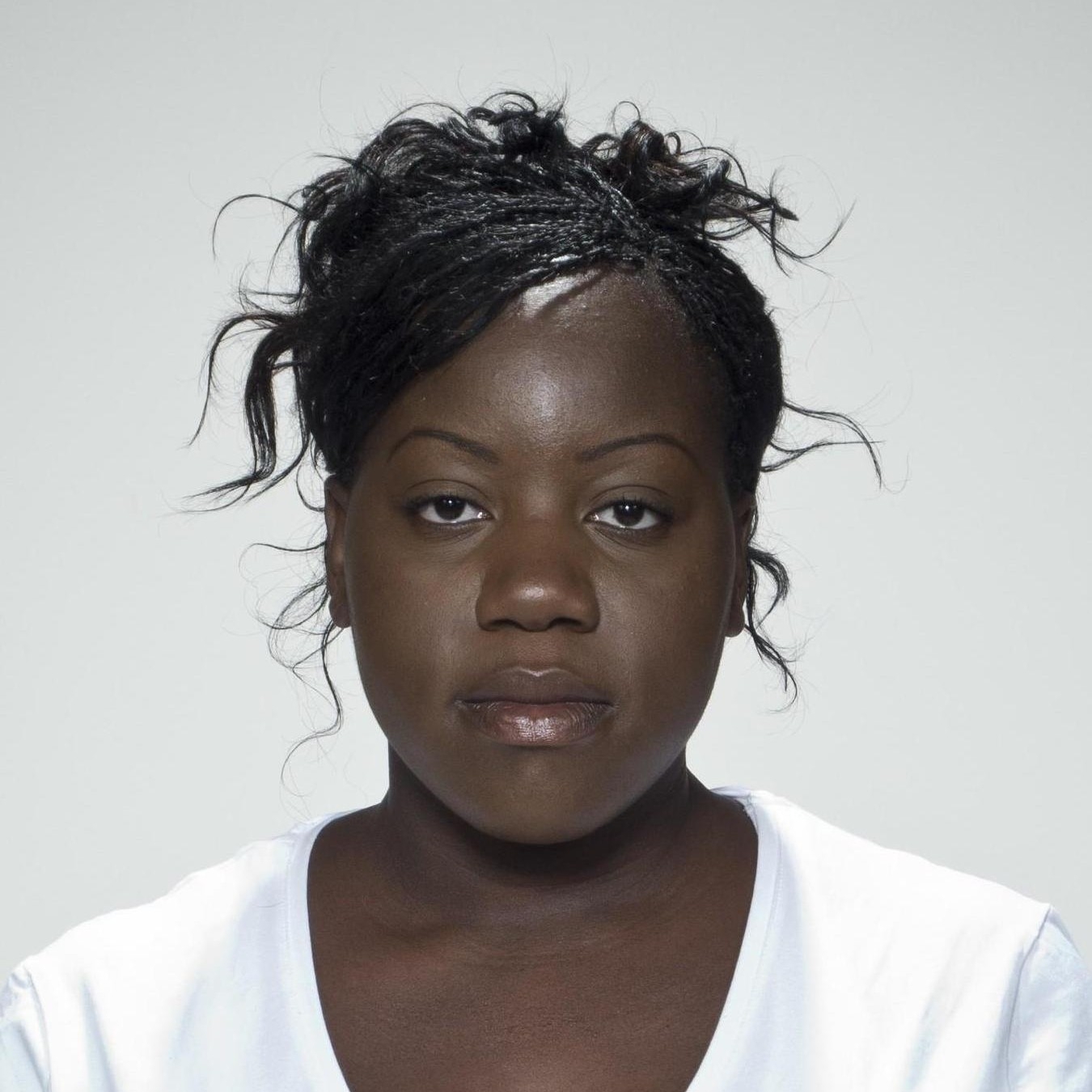}
         \caption{Subject 1}
         \label{fig:subject-1}
     \end{subfigure}
     \begin{subfigure}[b]{0.35\linewidth}
         \centering
         \includegraphics[width=\linewidth]{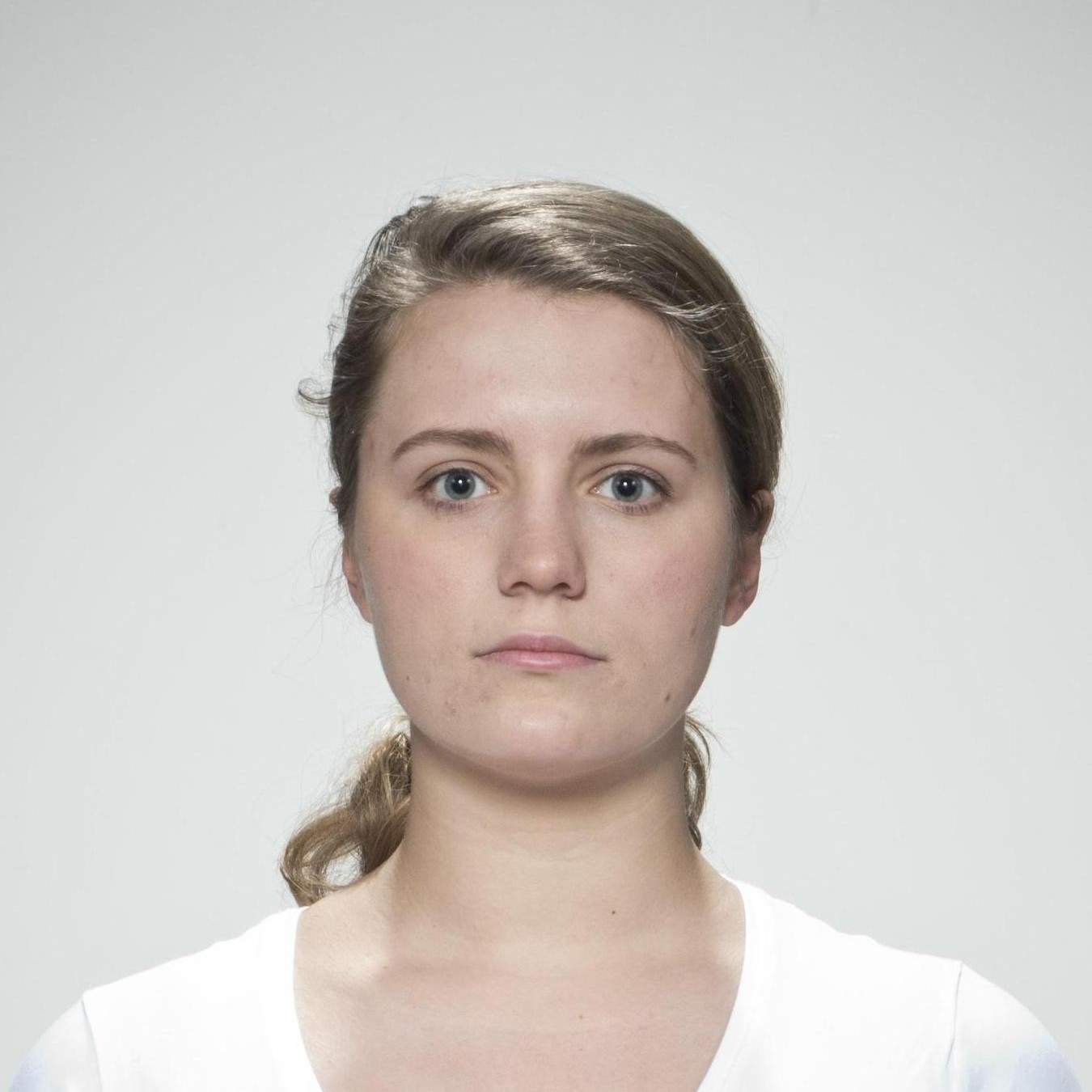}
         \caption{Subject 2}
         \label{fig:subject-2}
     \end{subfigure}
     \vspace{-0.25cm}
    \caption{Images from FRLL dataset \cite{b7}}
    \label{fig:subjects}
\end{figure}

A good FIQA is expected to be robust against multiple variations in the images they process including the consideration to demographic variations \cite{b16}.

FIQA algorithms depend on analyzing a face image to extract features and measure certain aspects of the image. They should be robust to numerous variations in the input images, herein demographic variations. For instance, in a recent report by Amnesty International \cite{b14}, it is reported that \textit{"Facial recognition systems misidentify Black faces at a high rate. Facial recognition is less accurate in identifying people with darker skin tones—especially women"}.  The ISO/IEC CD3 on 29794-5 standard \cite{b16} states that \textit{"A face image quality assessment algorithm should be insensitive to demographic factors such as age, skin-tone or ethnicity"}. For instance, given the two subjects in Figure \ref{fig:subjects} with two different skin tones, we would expect a FIQA algorithm that is assessing the illumination conditions, exposure, or natural color, to give relatively similar results given the images were taken under the same photo session setup made according to the ICAO requirements formulated in ISO/IEC 39794-5 in Annex D.1 \cite{b15}. Thus, when developing facial biometric systems or FIQA methods, rigorous testing should be performed to make sure the algorithms are robust to demographic differences to uncover any major differences in performance.

Face image quality component measures in ISO/IEC 39794-5 such as dynamic range, over- and under-exposure are measured on the skin of the face. These measures should be carefully designed to take into consideration variations in skin tones, and should not make assumptions about the true skin tone of the subject as estimating the true skin tone from an image is not a reliable process \cite{b13}. In 2021, Howard et al. \cite{b13} presented a study that explores the relationship between measured skin tone estimates from face images, and ground-truth skin readings collected using a colormeter device specifically designed to measure human skin. The study established that skin tones estimated from different images of the same subject varied significantly from the ground-truth skin tone values. The study also showed that estimated skin tone measures as part of a face recognition application lead to erroneous results depending on algorithms measuring some skin tone features across different skin tones, even for same subject under different environmental conditions. The analysis suggested that alternative methods that do not rely on the face skin should be used to avoid any noisy or biased results \cite{b13}.

One part of the face that has consistent color across all demographic groups is the sclera of the eye, whose whitish color has been shown to be a general characteristic of the human eye \cite{b17} \footnote{The sclera can become reddish in some unusual and medical situations. This happens when the small blood vessels on its surface become dilated or irritated due to various potential causes such as dry eyes, allergies, infections, Glaucoma, and Iritis. \cite{b19}}. In a recent work, Kabbani et al. \cite{b18} demonstrated that the behavior of the statistical features of the pixels of the eye sclera region is consistent across demographic groups. In this work, we examine the possibility of using the eye sclera for the assessment of dynamic range, over- and under-exposure components as an alternative FIQA approach which is both agnostic to skin tones and a fair FIQA. We present a comprehensive evaluation of dynamic range, over- and under-exposure on sclera and compare it against the same measures using the face image. We present a detailed analysis using EDC analysis to demonstrate sclera as an alternative candidate to estimate face image quality with similar performance while being skin-tone agnostic FIQA measures.

\section{Related Works}

The eye sclera has received attention in the literature primarily as a biometric recognition modality, not as a mean for face image quality assessment modality. It has been shown that sclera recognition, as one of the ocular traits, has high recognition accuracy and considerable user acceptance, and while iris recognition is the primary technology in the ocular biometrics group, sclera recognition, and particularly the vasculature of the sclera, has recently been considered as a complement or a substitute to iris recognition \cite{b25}. The vascular structure in the sclera region is unique for each individual and relatively stable over the subject's life time. Early in 2012, Zhou et al. \cite{b28} proposed to use sclera as a mean for uniquely identifying subjects. They introduce a method for sclera segmentation, and design a Gabor wavelet-based sclera pattern enhancement method to emphasize and binarize the sclera vessel patterns which are often blurry or have low contrast due to the highly reflective nature of the sclera area. They also propose a method to do feature extraction based on converting the vessel structure to a set of single-pixel wide lines, which are then used to create the subject's sclera template. These templates are later used for recognition. In 2021, Das et al. \cite{b6} propose a lightweight deep learning model based on UNet architecture for sclera segmentation, an unsupervised methods for vessel extraction as well as a gaze detection model. They introduce DeepR, a deep model for sclera recognition which compares two vessel-structure pairs and produces a boolean output on whether they match or not. The proposed methods are trained and tested over the MASD dataset consisting of 164 subjects, and the best reported results achieve 0.97 Area under the curve (AUC) in recognition accuracy.

FIQA measures are undergoing a standardization process in ISO/IEC 29794-5 standard \cite{b16}, and a reference implementation is under development in the Open Source Face Image Quality (OFIQ) project which will provide an open-source framework than can be deployed in commercial and government applications \footnote{https://github.com/BSI-OFIQ/OFIQ-Project}. A FIQA measure can be either an end-to-end unified quality score, or a quality component that addresses a specific component of the face image.

Many works in the literature proposed end-to-end unified quality scores. In 2020, Terhörst et al. \cite{b26} present an unsupervised approach for estimating a face image quality. The proposed method assesses the variations in the embeddings generated from random sub-networks of a face model. These variations in a sample representation are used to produce a unified score for a given face image. In 2021, Meng et al. \cite{b21} present another unified quality score that depends on the magnitude of the feature embedding of a face image. They introduce a new loss for training a face recognition model, and prove that the magnitude of the feature embedding, generated by the trained model, monotonically increases as the subject in a face image is more likely to be recognized. Later in 2021, Boutros et al. \cite{b1} present a method for unified FIQA that depends on measuring the relative sample classifiability which is measured based on the allocation of the training sample feature representation in angular space with respect to its class center and the nearest negative class center. They illustrate that there is a correlation between the face image quality and the sample relative classifiability.

Many other works addressed individual quality components. In 2021, Hernandez-Ortega \cite{b12} propose a framework to assess several quality components such as exposure, unnatural color, expression, pose, and background uniformity.The framework assesses image exposure by detecting if the image is too dark (or too bright) based on its mean pixel value. It uses a color-detector to detect pixels with unnatural color, and a pretrained CNN to detect the facial expression. In 2022, Yin and Chen \cite{b27} introduce a face segmentation model that can be used to detect the occluded parts of the face. In 2023, Grimmer et al. \cite{b10} proposes an expression neutrality quality assessment method, NeutrEx, that is based on the accumulated distances of a 3D face reconstruction to a neutral expression anchor \cite{b4} \cite{b8}. Many more methods exist for other quality components such as assessing eyes openness \footnote{https://github.com/sahnimanas/Fatigue-Detection} \footnote{https://github.com/zykev/eye\_state}, and mouth closedness \footnote{https://github.com/mauckc/mouth-open}.

Bias in FIQA is a less explored topic in the literature. A single work is found by Babnik and Štruc \cite{b3} on bias in FIQA with regard to end-to-end methods that produce a unified quality score. They assess demographic bias in terms of two demographic factors, sex and race, on a variety of quality assessment methods. They find that general-purpose image quality assessment methods, called no-reference IQA, specifically, BRISQUE \cite{b22}, NIQE \cite{b23} and RankIQA \cite{b20}, are less biased with respect to the two demographic factors considered, and that dedicated face image quality assessment methods, specifically, SDD-FIQA \cite{b24}, MagFace \cite{b21}, CR-FIQA \cite{b1}, and SER-FIQ \cite{b26} have strong bias with a tendency to favor white individuals of either sex.

\section{Methods}
In this section, we present the details on dataset, pre-processing and evaluation approaches.
\textbf{Dataset:} The dataset used for the experiments is the Face Research Lab London dataset (FRLL) \cite{b7}. The dataset contains 1020 full color images of 1350x1350 pixels for 102 subjects. It contains five pose variations per subject (frontal, left profile, right profile, left 3 quarter, right 3 quarter) and two expression variations (neutral and smiling). We use frontal images of each subject in this work to avoid an overlay of impact from strong pose variations on the recognition performance. Further, segmenting the sclera region of both eyes is not possible with the profile views for our analysis. 

\begin{figure}[h!]
     \centering
     \begin{subfigure}[b]{0.48\linewidth}
         \centering
         \includegraphics[width=0.48\linewidth]{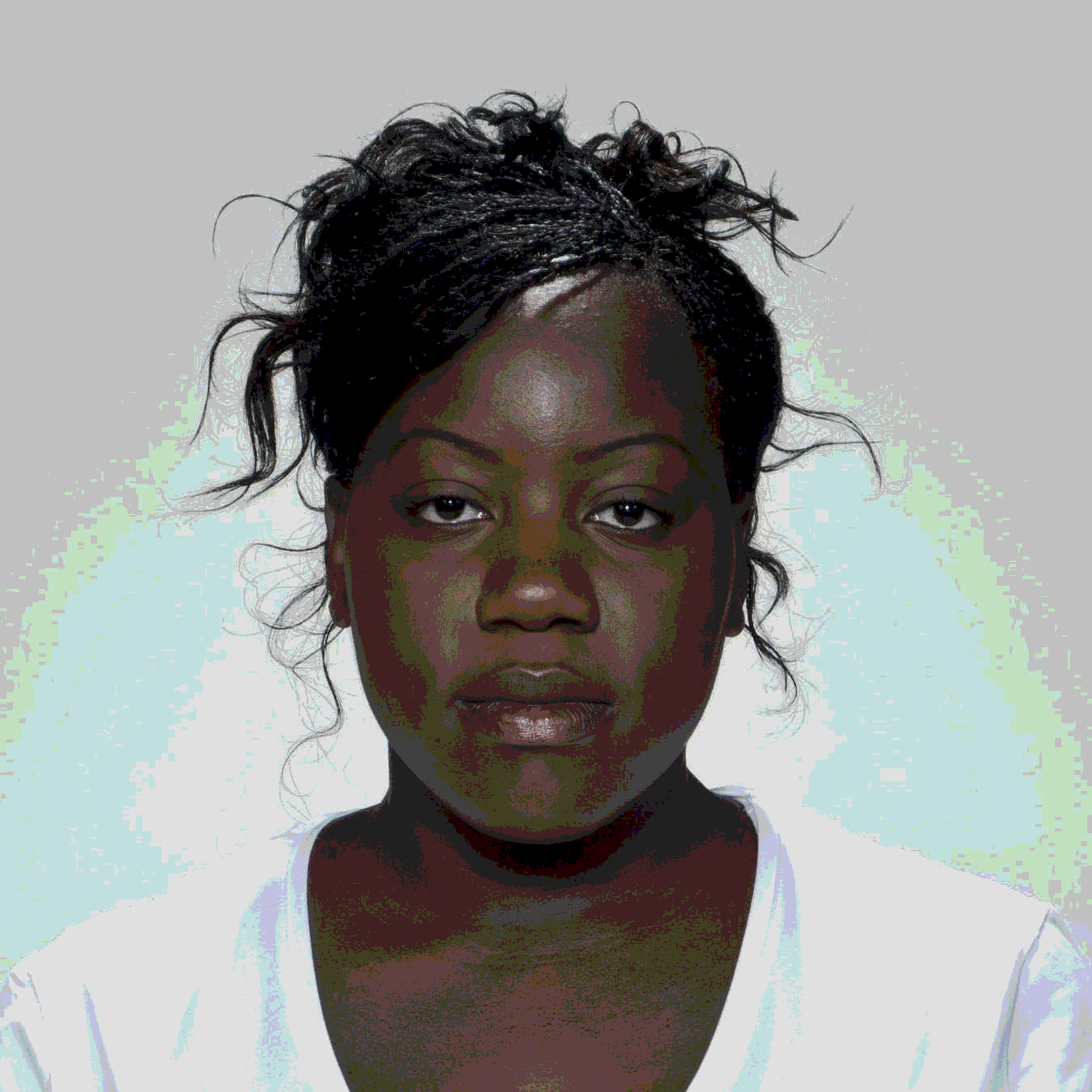}
         \includegraphics[width=0.48\linewidth]{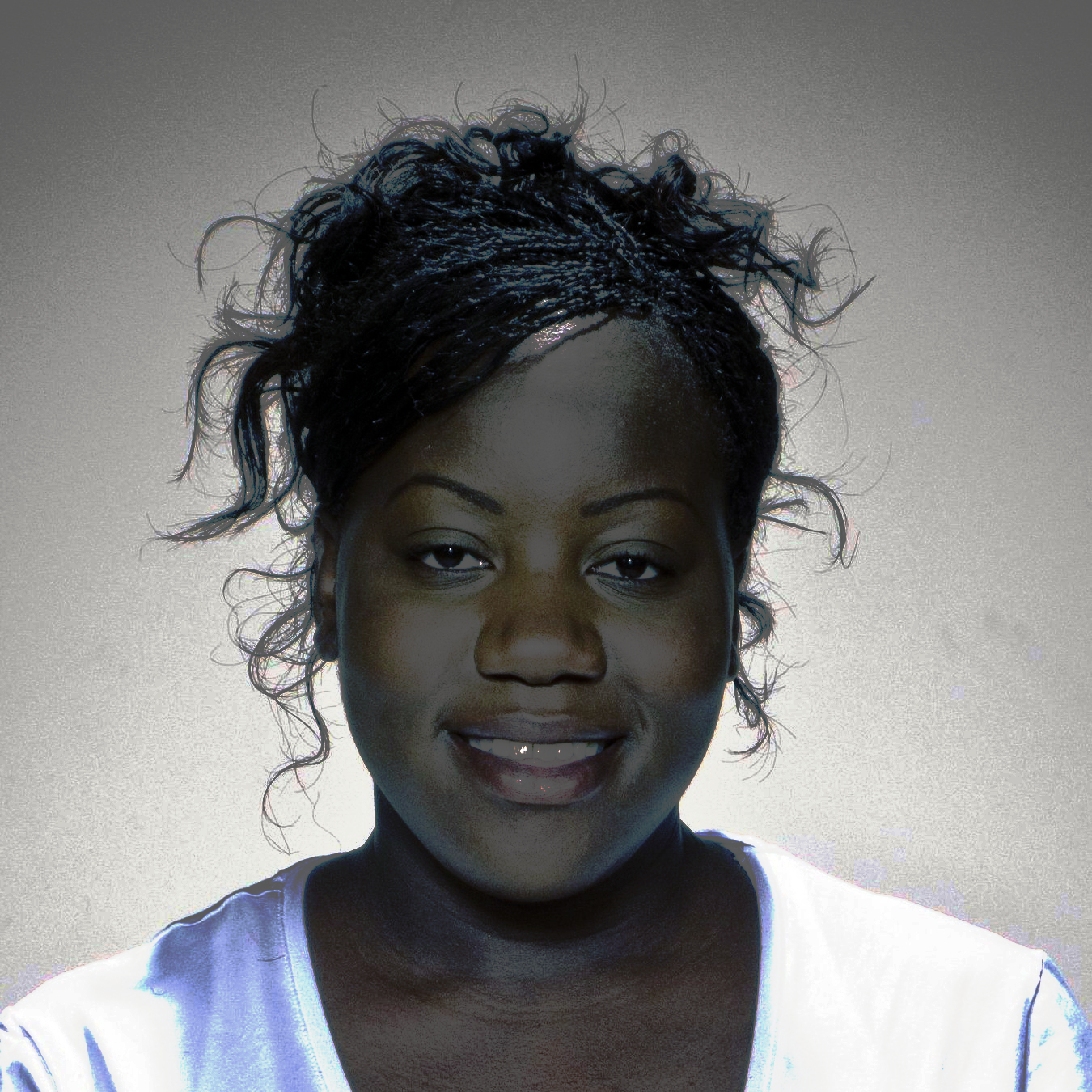}
         \caption{Subject 1}
         \label{fig:dynamic-range-s1}
     \end{subfigure}
     \hfill
     \begin{subfigure}[b]{0.48\linewidth}
         \centering
         \includegraphics[width=0.48\linewidth]{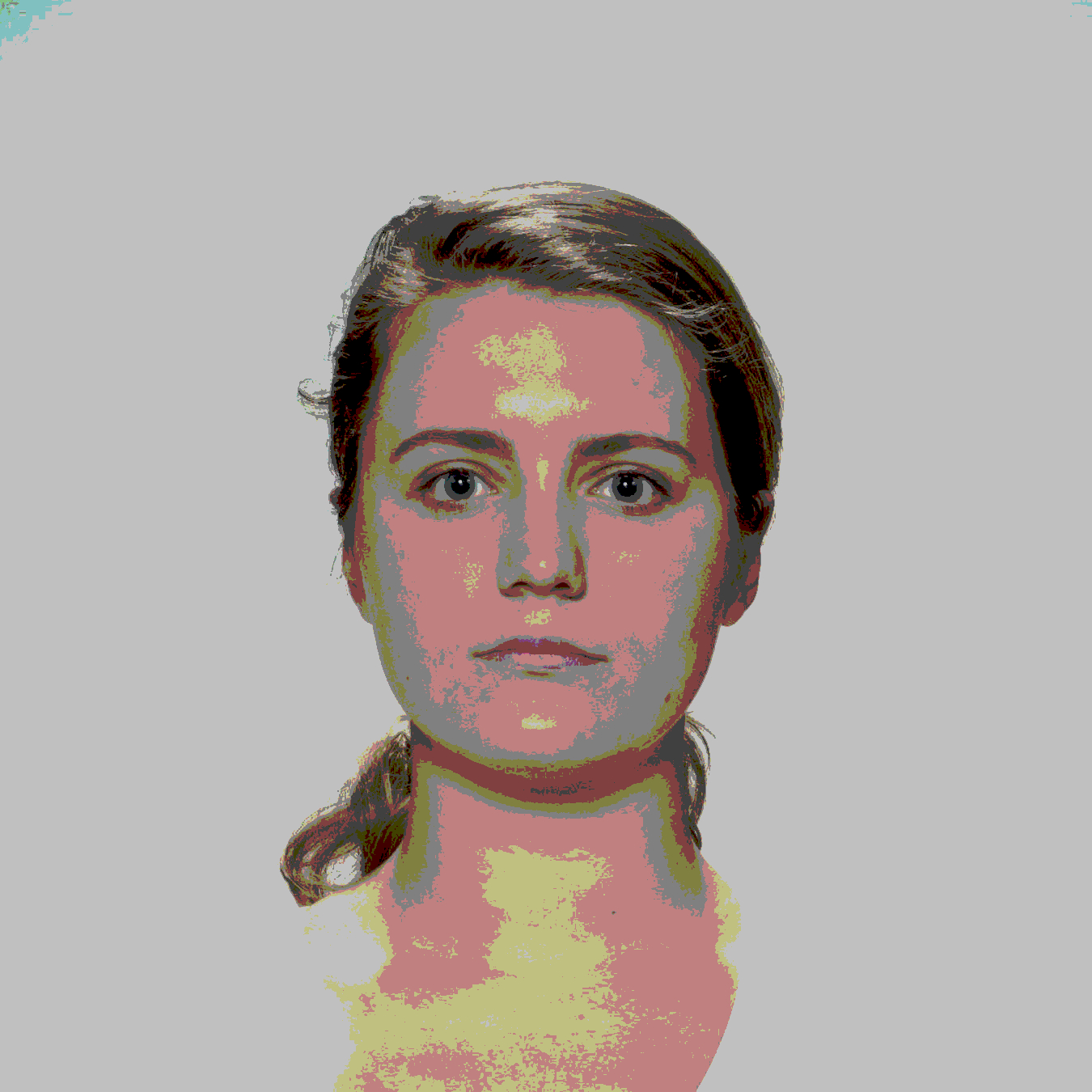}
         \includegraphics[width=0.48\linewidth]{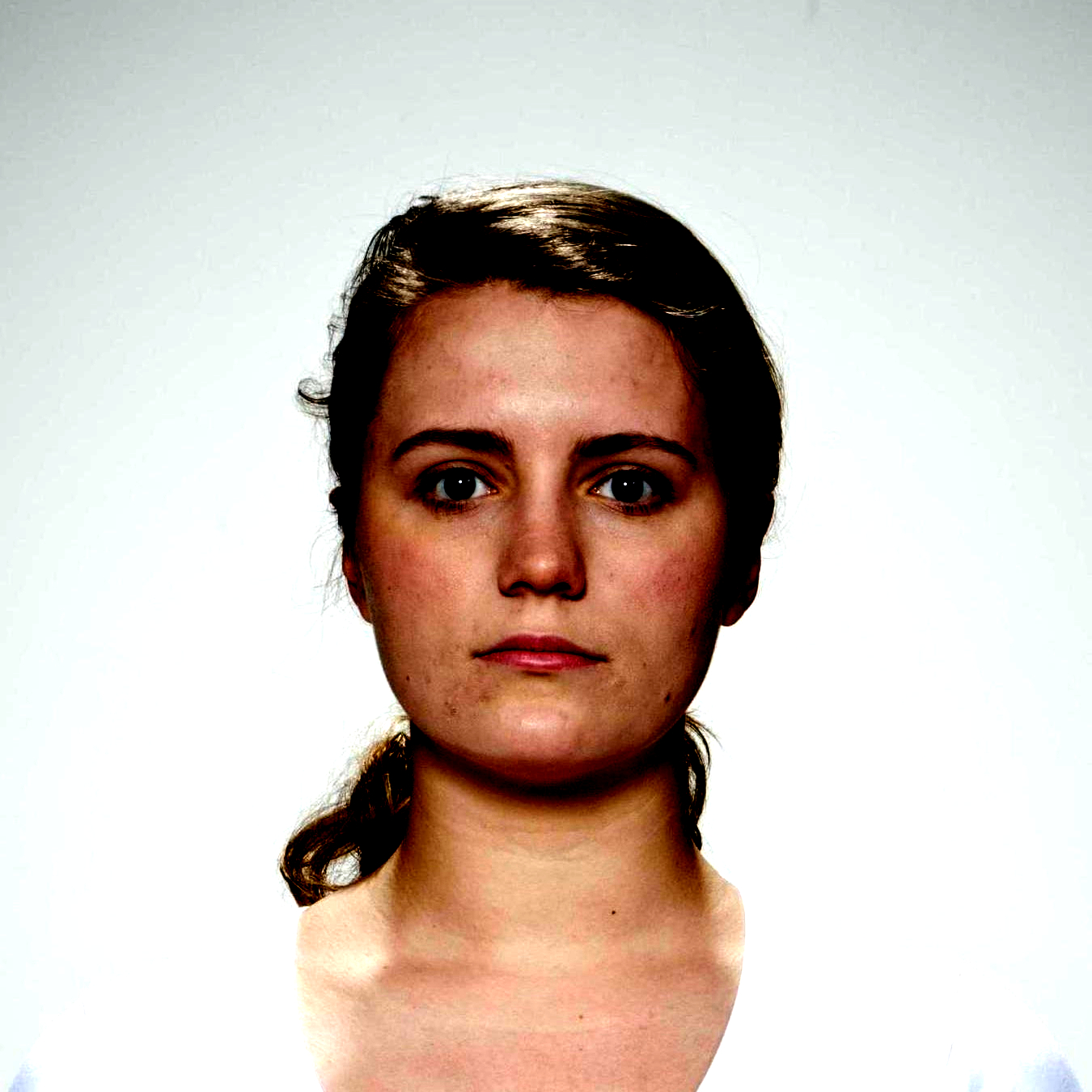}
         \caption{Subject 2}
         \label{fig:dynamic-range-s2}
     \end{subfigure}
    \caption{Examples of non-compliant images due to low dynamic range}
    \label{fig:dynamic-range-samples}
\end{figure}
To be able to analyze and evaluate the FIQA algorithms on demographic groups with different skin tones separately, the dataset is divided into two subsets. The first subset contains subjects of darker skin tone, we call it D-FRLL, and contains in total 17 subjects (9 males, 8 females). The second subset contains subjects of lighter skin tone, we call it L-FRLL, and contains in total 57 subjects (24 males, 33 females). The rest of the subjects were excluded either because they do not strictly belong to either of the groups, or because of the presence of a heavy beard for some male subjects. The subjects in each of the subsets were selected based on manual visual inspection. The FRLL is not a balanced dataset in terms of the number of subjects in each skin tone category, but it is a good choice for our analysis because the photos are taken in exactly the same photo session setup, with the same lighting, pose and expression conditions, making skin tone the only factor of variation.

\textbf{Augmentations}: Evaluating the FIQA on D-FRLL and L-FRLL will not produce representative results for the algorithms as there is not enough variations in dynamic range or exposure. Thus, to evaluate the FIQA algorithms and to examine differences in performance between using the face skin vs. using the sclera, the two subsets are created with additional images by introducing systematic synthetic variations in dynamic range and exposure.

\textbf{Preprocessing and skin extraction}: After augmentation, all datasets are processed such that the face is detected, aligned, and the image is cropped to the face region only. Sample and Computation Redistribution for Efficient Face Detection (SCRFD) is used as a face detection method \cite{b11}.  The face skin is extracted using the face parsing network \footnote{https://github.com/zllrunning/face-parsing.PyTorch} standardized in ISO/IEC CD 29794-5 \cite{b16} and 19 different areas such as hair, eyeglasses, eyes, eyebrows, nose, mouth and ears are segmented. The sclera is extracted using the landmark based method introduced in \cite{b18}.

\textbf{Face Recognition and FIQA}: ArcFace \cite{b5} is used as face recognition method for measuring the FRS performance.  "Error versus Discard Characteristic" (EDC)\cite{b25} are used to analyze the quality of biometric sample and to report partial Area Under Curve (pAUC).

\section{Dynamic Range}

The ISO/IEC 39794-5 standard \cite{b15} specifies the following requirement with regard to dynamic range. \textit{"The dynamic range of the image should have at least 50\% of intensity variation in the face region of the image"}. Figure \ref{fig:dynamic-range-samples} shows examples of face images with non-compliant low dynamic range.

The dynamic range assessment algorithm in this work is adapted from ISO/IEC CD3 29794-5 \cite{b16} and illustrated in Algorithm \ref{alg:dynamic}. Given face skin or sclera RGB data, it uses the the luminance histogram of the input pixels, and computes its entropy to produce an assessment value which is then mapped to the range of [0-100] using a sigmoid function. Unlike the algorithm described in the ISO/IEC CD3 29794-5 standard, which takes as input a face image including the eyes and the eyebrows, and uses $12.5 * H$ as the mapping function, this one takes as input either the sclera region or the face skin without eyes, eyebrows, lips, or teeth (if visible). It also uses a sigmoid mapping function to obtain consistent output values across different type of input.

\begin{algorithm}[h!]
\caption{Dynamic range quality assessment}
\label{alg:dynamic}
\KwData{face skin or sclera RGB data}
\KwResult{dynamic range quality component value}
$l \gets luminance(rgb)$\;
$h_0,h_1,\cdots,h_M \gets histogram(l)$\;
$H \gets \sum_{i=0}^{M}h_i log_2(h_i)$\;
$output \gets round(100 * sigmoid(H, 5, 1))$\;
\end{algorithm}

To evaluate the algorithm, and given that the FRLL dataset does not contain images with dynamic range variations. Augmentations are applied to the images in the evaluation datasets to synthetically produce various degrees of low dynamic range images. The new evaluation datasets will be called DR-D-FRLL, DR-L-FRLL, DR-FRLL for the dynamic range augmented datasets D-FRLL, L-FRLL, and FRLL respectively.

\begin{figure}[h!]
     \centering
     \begin{subfigure}[b]{0.3\linewidth}
         \centering
         \includegraphics[width=\linewidth]{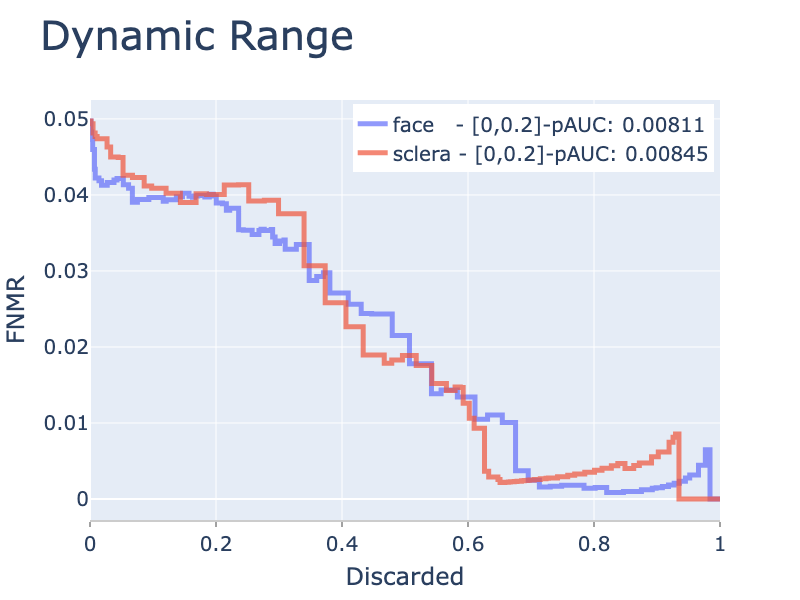}
         \caption{All}
         \label{fig:edc-dr-all}
     \end{subfigure}
     \begin{subfigure}[b]{0.3\linewidth}
         \centering
         \includegraphics[width=\linewidth]{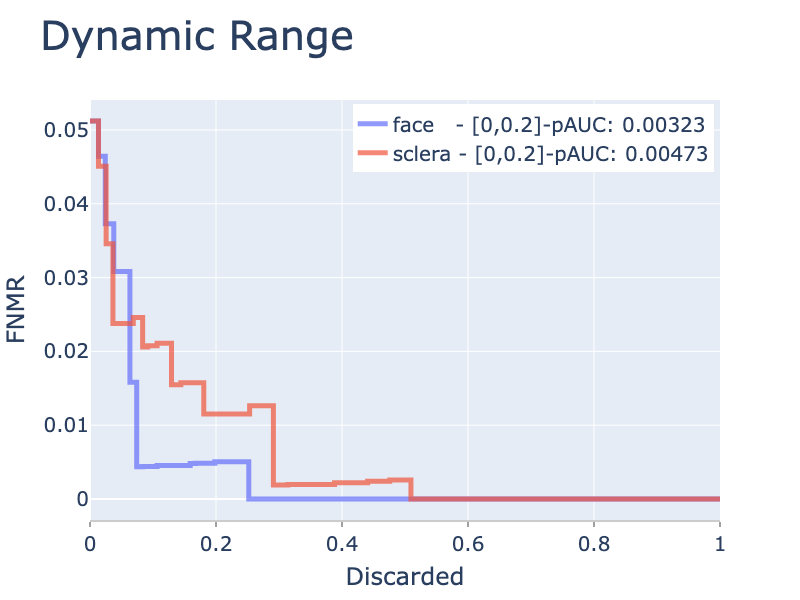}
         \caption{Dark}
         \label{fig:edc-dr-dark}
     \end{subfigure}
     \begin{subfigure}[b]{0.3\linewidth}
         \centering
         \includegraphics[width=\linewidth]{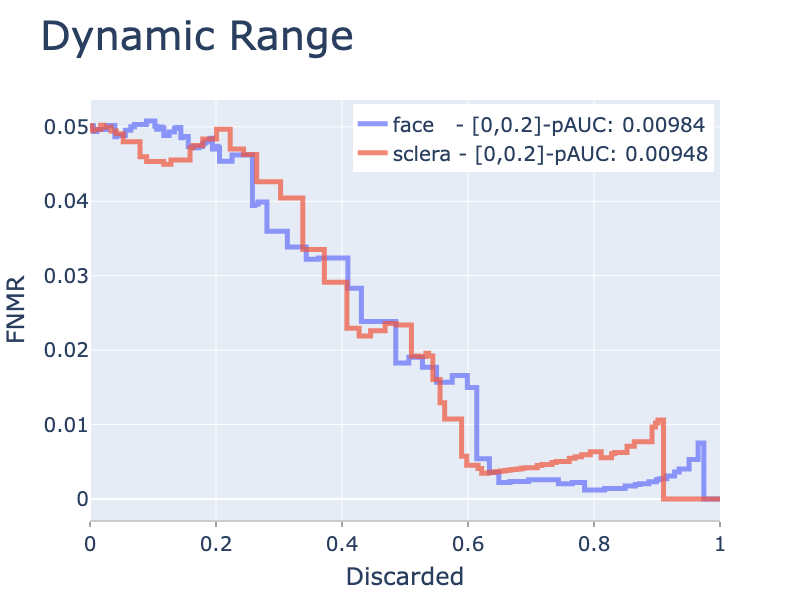}
         \caption{Light}
         \label{fig:edc-dr-light}
     \end{subfigure}
        \caption{Dynamic range EDC curves for face vs. sclera with a starting error rate of 0.05 and ArcFace as face recognition system.}
        \label{fig:edcs-dynamic-range}
\end{figure}

Figure \ref{fig:edcs-dynamic-range} shows the evaluation results of the dynamic range assessment algorithm (Alg~\ref{alg:dynamic}) on the DR-D-FRLL subset in Figure \ref{fig:edc-dr-dark}, the DR-L-FRLL subset in Figure \ref{fig:edc-dr-light}, and the combined dataset DR-FRLL in Figure \ref{fig:edc-dr-all}. The EDC curves show that the algorithm has similar performance whether the input is the face skin or the sclera. This means that the predictive capacity of both the face skin and the sclera with regard to measuring dynamic range is the same. Furthermore, dynamic range seems to have a strong impact on the face recognition performance given how the curve is descending consistently as more lower quality images are discarded. It is also noticeable that dynamic range has less impact on the face recognition performance for people with darker skin tone given that the error rate descends rapidly as a small portion of images is discarded, while the error rate falls at a much slower pace as lower quality images are discarded for the lighter skin tone individuals.

\begin{figure}[h!]
     \centering
     \begin{subfigure}[b]{0.45\linewidth}
         \centering
         \includegraphics[width=\linewidth]{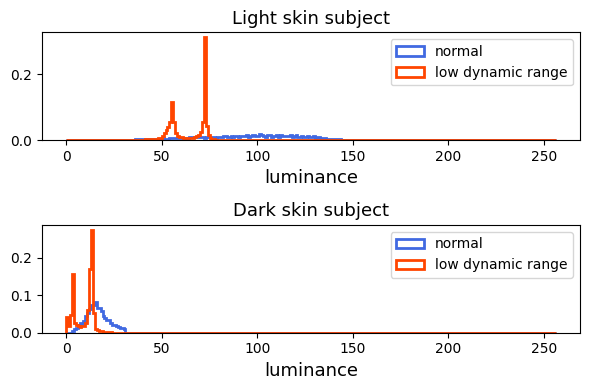}
         \vspace{-0.45cm}
         \caption{Face}
         \label{fig:dr-lumi-face}
     \end{subfigure}
     \begin{subfigure}[b]{0.45\linewidth}
         \centering
         \includegraphics[width=\linewidth]{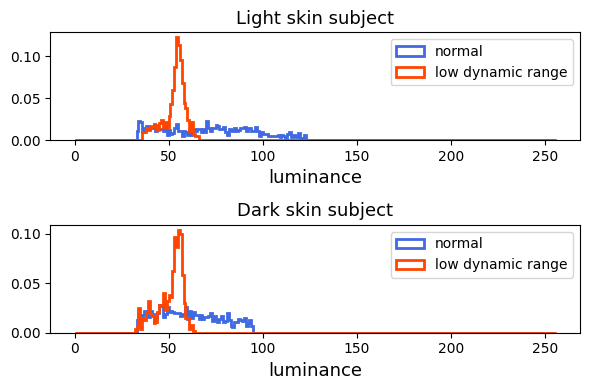}
         \vspace{-0.45cm}
         \caption{Sclera}
         \label{fig:dr-lumi-sclera}
     \end{subfigure}
     \vspace{-0.25cm}
    \caption{Normal vs. Low dynamic range for face skin vs. sclera}
    \label{fig:dr-lumi}
\end{figure}

Figure \ref{fig:dr-lumi} shows the luminance histograms of normal vs. low dynamic range, and face skin vs. sclera images of two sample subjects. Figure \ref{fig:dr-lumi-face} on the left shows the histograms for the face skin and Figure \ref{fig:dr-lumi-sclera} on the right shows the ones for the sclera. It is very clear in both cases how the low dynamic range images have spiky histograms with less spread values since the bit range of the individual pixels is limited compared to normal images. It is also clear that the histograms for both the normal and the low dynamic range images differ between light and dark skin tone subjects, while they are much more similar for the sclera histograms. Column 3 in Table \ref{tbl:distances} shows that the distance between the two sclera histograms is much smaller than the distance between the two face skin histograms of the low dynamic range images on two distance measures.

\begin{table}[]
\centering
\resizebox{\linewidth}{!}{%
\begin{tabular}{|l|ll|ll|ll|ll|}
\hline
& \multicolumn{2}{l|}{Over-exposure} & \multicolumn{2}{l|}{Under-exposure} & \multicolumn{2}{l|}{Low Dynamic Range} \\ \hline
Measure & \multicolumn{1}{l|}{Sclera} & Face & \multicolumn{1}{l|}{Sclera} & Face & \multicolumn{1}{l|}{Sclera} & Face \\ \hline
Chi-Squared Distance & \multicolumn{1}{l|}{0.09}  & 0.60  & \multicolumn{1}{l|}{0.05}  & 10.15   & \multicolumn{1}{l|}{0.02}    & 0.20    \\ \hline
Hellinger Distance & \multicolumn{1}{l|}{0.19}   & 0.97 & \multicolumn{1}{l|}{0.54}   & 0.87 & \multicolumn{1}{l|}{0.18}   & 1.0  \\ \hline
\end{tabular}%
}
\caption{Comparison of the distances between the sclera histograms and the face skin histograms of subject 1 and 2 in three types of image conditions and on two distance measures. It is clear that under the three image conditions, the distance between the sclera histograms for light and dark skin tone subjects is always much smaller than the corresponding face skin distance.}
\label{tbl:distances}
\end{table}

\section{Under and Over Exposure}

\begin{figure}[h!]
     \centering
     \begin{subfigure}[b]{0.48\linewidth}
         \centering
         \includegraphics[width=0.48\linewidth]{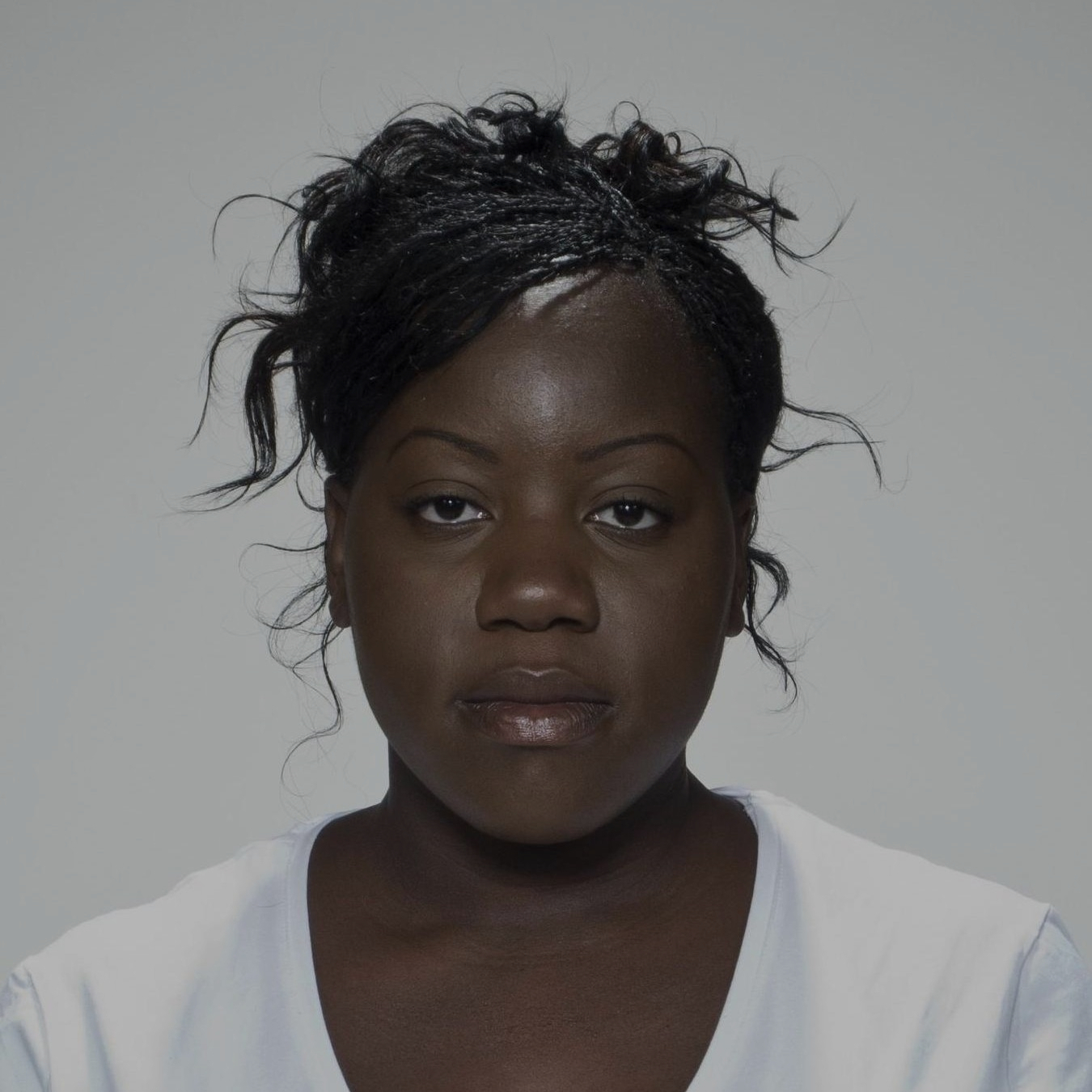}
         \includegraphics[width=0.48\linewidth]{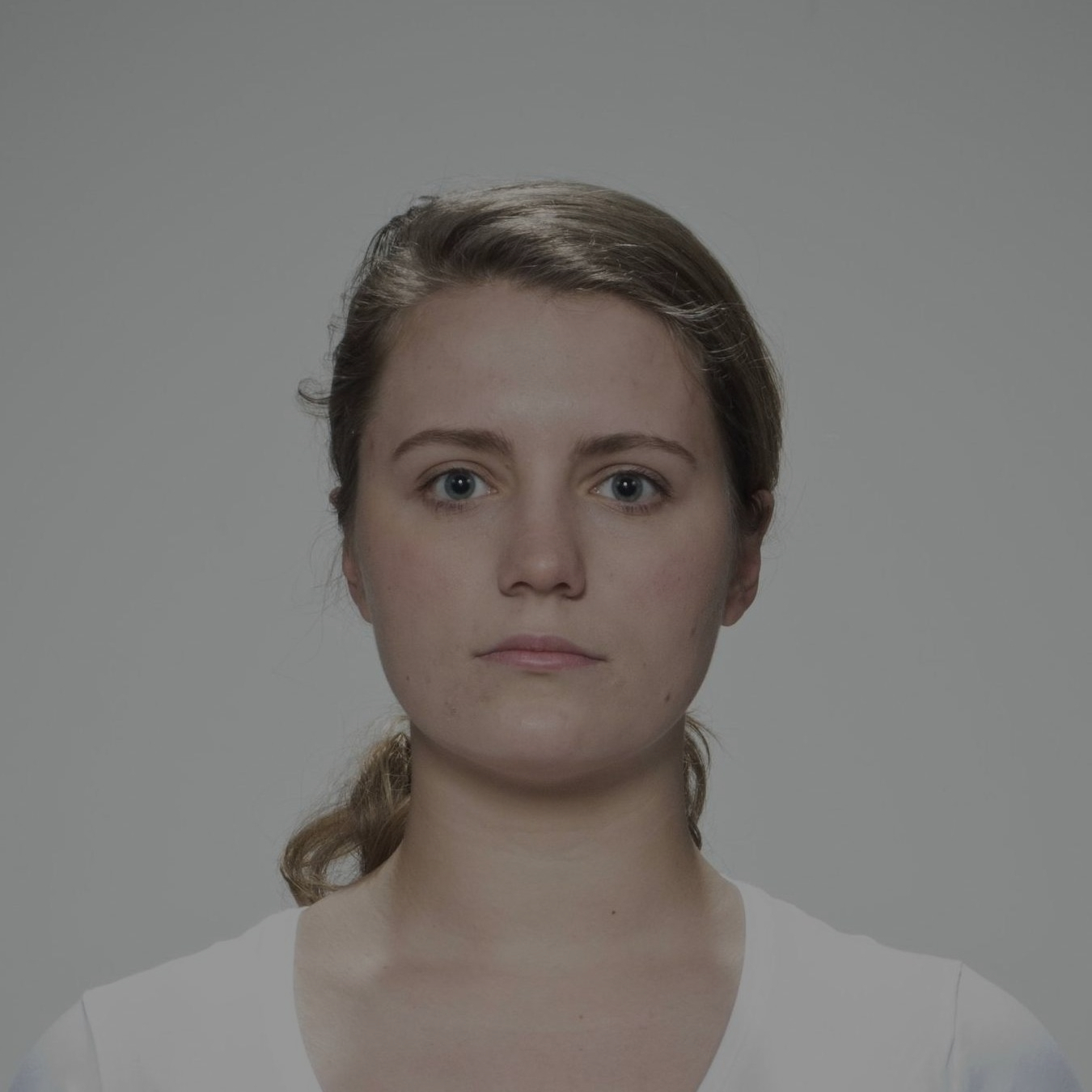}
         \caption{Under exposure}
         \label{fig:under-exposure-samples}
     \end{subfigure}
     \hfill
     \begin{subfigure}[b]{0.48\linewidth}
         \centering
         \includegraphics[width=0.48\linewidth]{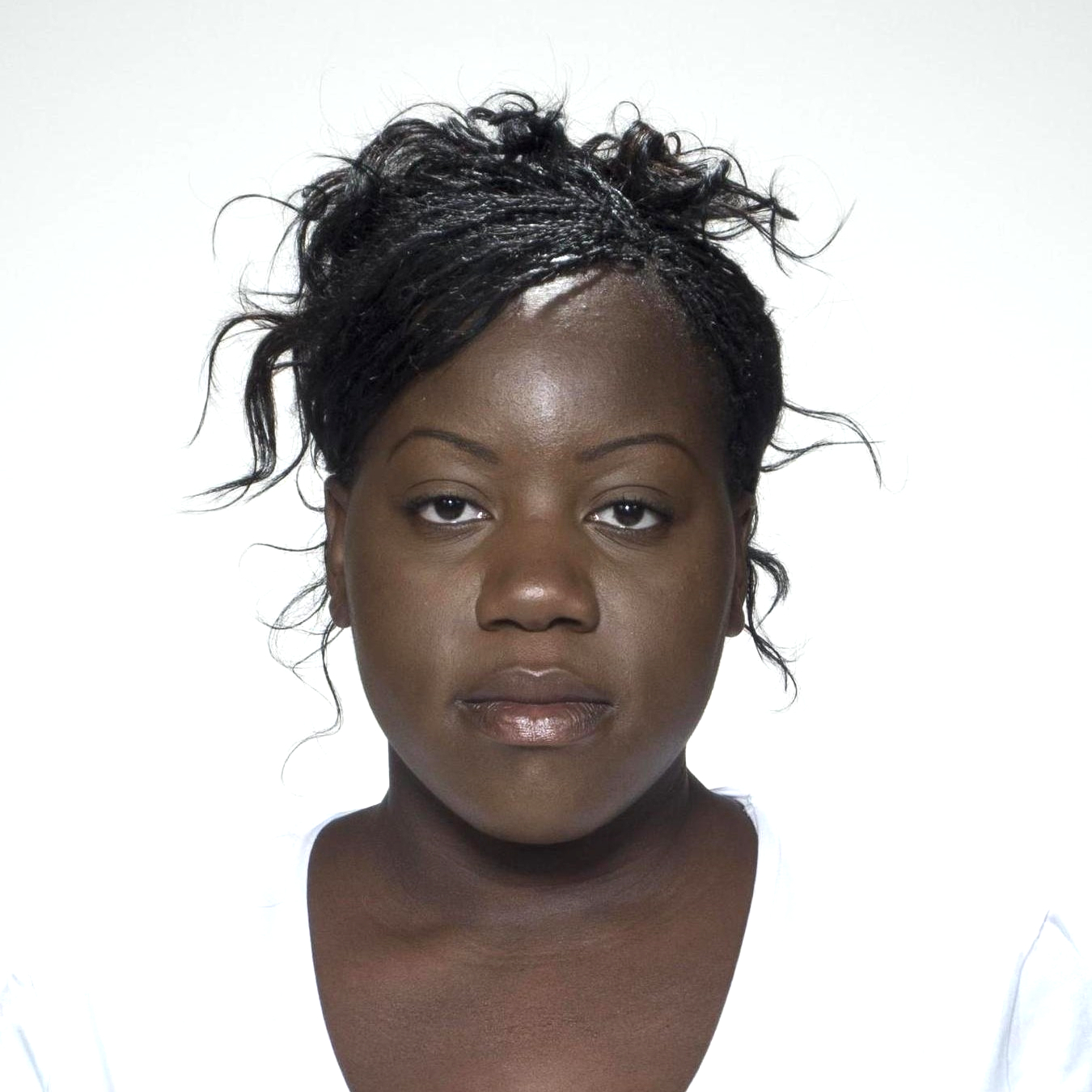}
         \includegraphics[width=0.48\linewidth]{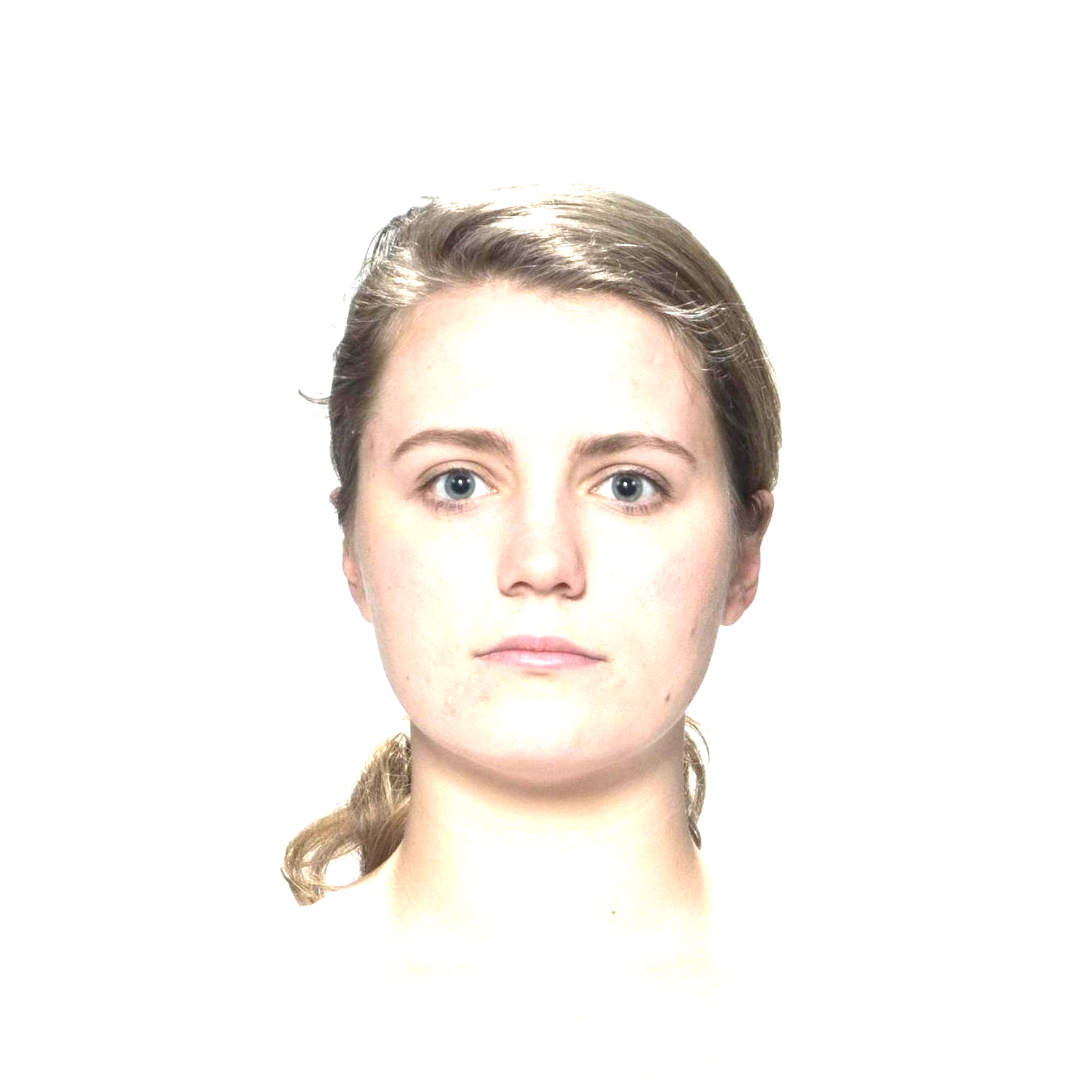}
         \caption{Over exposure}
         \label{fig:over-exposure-samples}
     \end{subfigure}
     \vspace{-0.25cm}
    \caption{Examples of non-compliant exposure}
    \label{fig:exposure-samples}
\end{figure}

The ISO/IEC 39794-5 standard \cite{b15} specifies the following requirement with regard to exposure. \textit{"The face portrait shall have appropriate brightness and good contrast between face, hair and background"}. Figure \ref{fig:exposure-samples} shows examples of face images with non compliant under and over exposure.

The under exposure assessment algorithm in this work is the same as the one described in ISO/IEC CD3 29794-5 \cite{b16} adapted to accept the face skin or the sclera as input. The algorithm is illustrated in Algorithm \ref{alg:under}. Given skin face or sclera RGB data, it uses the luminance histogram to produce an assessment value based on the number of low luminance values which is then mapped to the range of [0-100].

\begin{algorithm}[h!]
\caption{Under exposure quality assessment}
\label{alg:under}
\KwData{face skin or sclera RGB data}
\KwResult{under exposure quality component value}
$l \gets luminance(rgb)$\;
$h_0,h_1,\cdots,h_M \gets histogram(l)$\;
$v \gets \sum_{i=0}^{25}h_i$\;
$output \gets round(100 * (1 - v))$\;
\end{algorithm}

\begin{figure}[h!]
     \centering
     \begin{subfigure}[b]{0.3\linewidth}
         \centering
         \includegraphics[width=\linewidth]{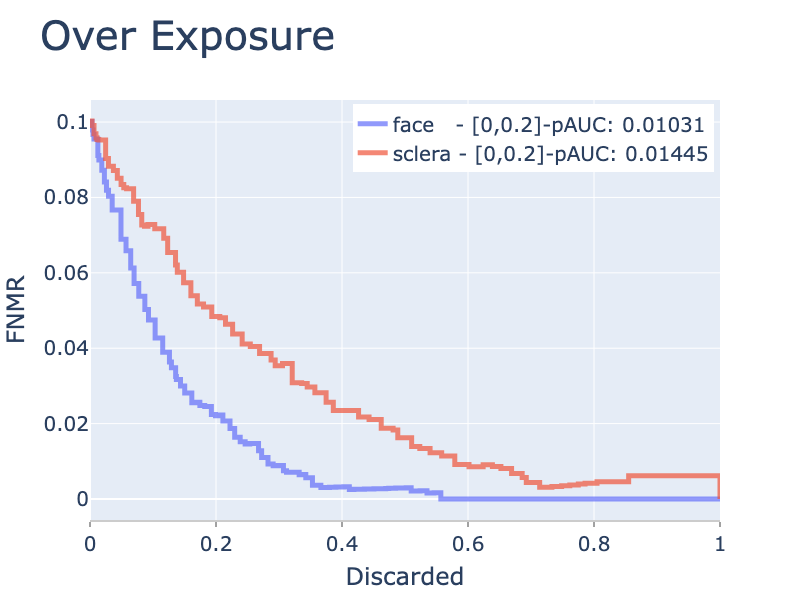}
         \caption{All}
         \label{fig:edc-ox-all}
     \end{subfigure}
     \begin{subfigure}[b]{0.3\linewidth}
         \centering
         \includegraphics[width=\linewidth]{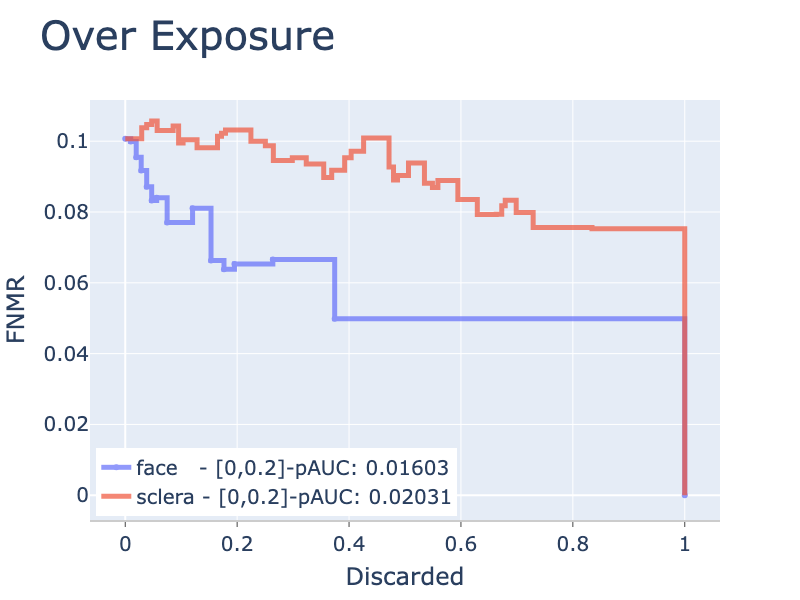}
         \caption{Dark}
         \label{fig:edc-ox-dark}
     \end{subfigure}
     \begin{subfigure}[b]{0.3\linewidth}
         \centering
         \includegraphics[width=\linewidth]{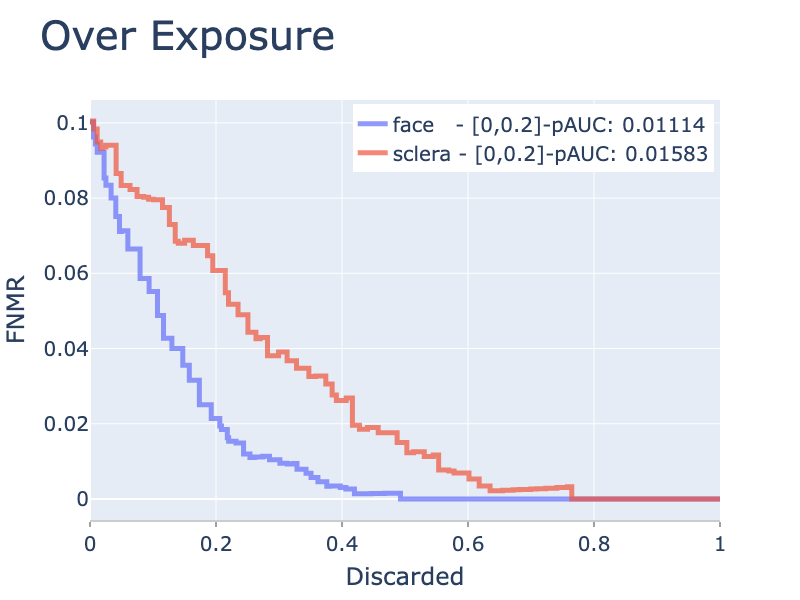}
         \caption{Light}
         \label{fig:edc-ox-light}
     \end{subfigure}
        \caption{Over exposure EDC curves for face vs. sclera with a starting error rate of 0.1 and ArcFace as face recognition system.}
        \label{fig:edc-ox}
\end{figure}

The over exposure assessment algorithm is also the same as the one described in ISO/IEC CD3 29794-5 \cite{b16} but adapted to accept the face skin or the sclera as input and with a different mapping function in the last step. It uses the same mapping function as the one for the under exposure algorithm because it was found to give better output values in terms of the distribution of values in the range [0,100]. The algorithm is illustrated in Algorithm \ref{alg:over}. Given skin face or sclera RGB data, it uses the luminance histogram to produce an assessment value based on the number of high luminance values which is then mapped to the range of [0-100].

\begin{algorithm}[h!]
\caption{Over exposure quality assessment}
\label{alg:over}
\KwData{face skin or sclera RGB data}
\KwResult{over exposure quality component value}
$l \gets luminance(rgb)$\;
$h_0,h_1,\cdots,h_M \gets histogram(l)$\;
$v \gets \sum_{i=247}^{255}h_i$\;
$output \gets round(100 * (1 - v))$\;
\end{algorithm}

To evaluate both algorithms, and given that the FRLL dataset does not contain images with different exposure conditions, augmentations are applied to the images in the evaluation datasets to synthetically produce under and over exposed images. The new evaluation datasets will be called EX-D-FRLL, EX-L-FRLL, EX-FRLL for the exposure augmented datasets D-FRLL, L-FRLL, and FRLL respectively.

Figure \ref{fig:edc-ox} shows the evaluation results of the over exposure assessment algorithm (Alg~\ref{alg:over}) on the EX-D-FRLL subset in Figure \ref{fig:edc-ox-dark}, the EX-L-FRLL subset in Figure \ref{fig:edc-ox-light}, and the combined dataset EX-FRLL in Figure \ref{fig:edc-ox-all}. The EDC curves show that the algorithm has similar performance whether the input is the face skin or the sclera wih slight advantage to the face skin. This means that the predictive capacity of both the face skin and the sclera with regard to over exposure is the same. Furthermore, it shows that over exposure has more impact on the darker skin tone as the face recognition error rate does not fall in the same way as for lighter skin tone subjects even as the portion of discarded lower quality images increases.

\begin{figure}[h!]
     \centering
     \begin{subfigure}[b]{0.3\linewidth}
         \centering
         \includegraphics[width=\linewidth]{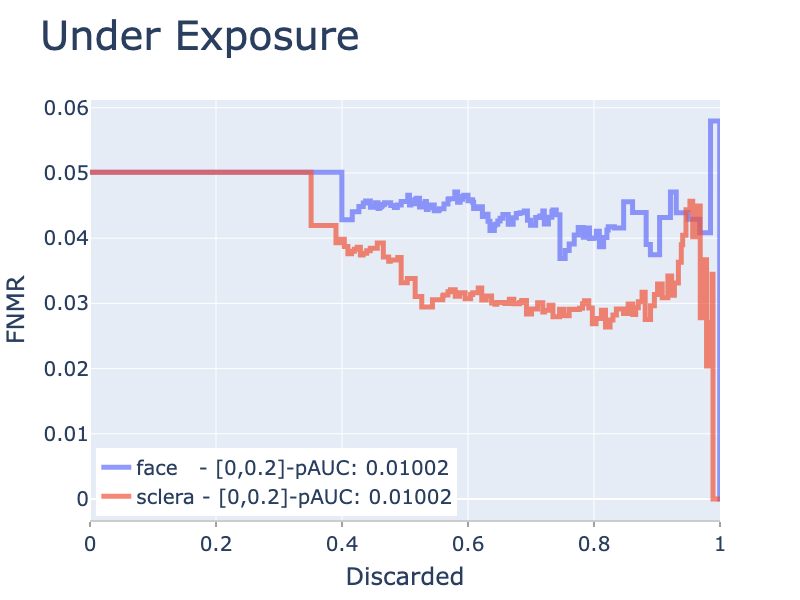}
         \caption{All}
         \label{fig:edc-ux-all}
     \end{subfigure}
     \begin{subfigure}[b]{0.3\linewidth}
         \centering
         \includegraphics[width=\linewidth]{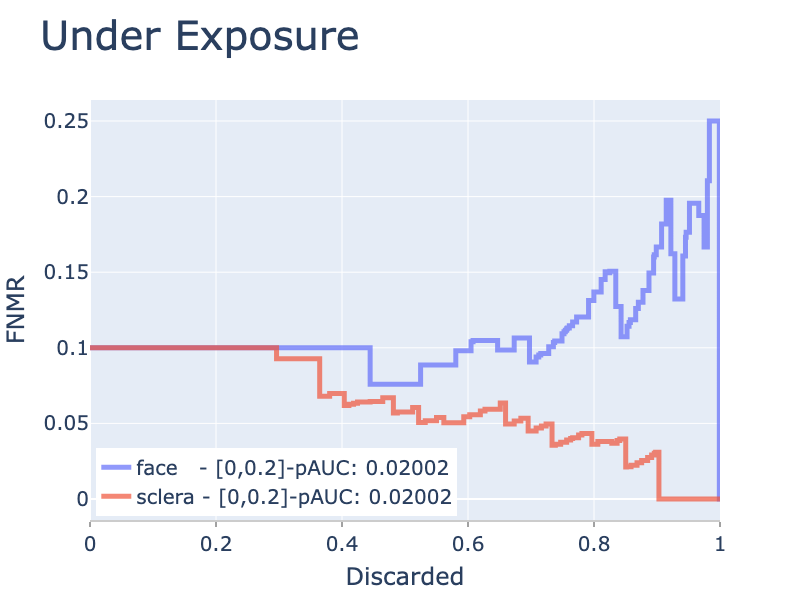}
         \caption{Dark}
         \label{fig:edc-ux-dark}
     \end{subfigure}
     \begin{subfigure}[b]{0.3\linewidth}
         \centering
         \includegraphics[width=\linewidth]{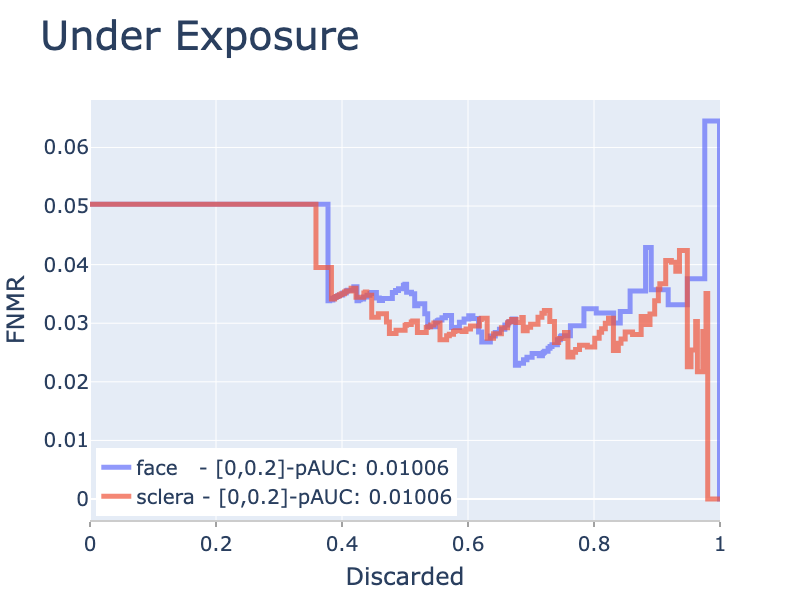}
         \caption{Light}
         \label{fig:edc-ux-light}
     \end{subfigure}
        \caption{Under exposure EDC curves for face vs. sclera with a starting error rate of 0.05 and ArcFace as face recognition system.}
        \label{fig:edc-ux}
\end{figure}

Figure \ref{fig:edc-ux} shows the evaluation results of the under exposure assessment algorithm (Alg~\ref{alg:under}) on the EX-D-FRLL subset in Figure \ref{fig:edc-ux-dark}, the EX-L-FRLL subset in Figure \ref{fig:edc-ux-light}, and the combined dataset EX-FRLL in Figure \ref{fig:edc-ux-all}. The EDC curves show that the algorithm has similar performance whether the input is the face skin or the sclera. This means that the predictive capacity of both the face skin and the sclera with regard to under exposure is the same. Furthermore, it shows that under exposure has a very strong impact on the face recognition performance because the error rate does not fall considerably even after a large portion of the lowest quality images are discarded, and the impact is more or less the same for darker and lighter skin tone subjects.

\begin{figure}[htp]
     \centering
     \begin{subfigure}[b]{0.45\linewidth}
         \centering
         \includegraphics[width=\linewidth]{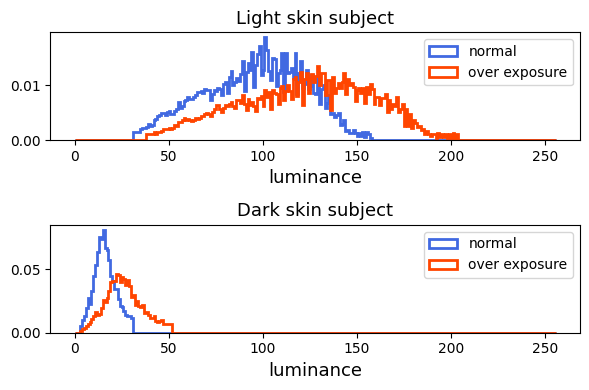}
         \vspace{-0.45cm}
         \caption{Face}
         \label{fig:ox-lumi-face}
     \end{subfigure}
     \begin{subfigure}[b]{0.45\linewidth}
         \centering
         \includegraphics[width=\linewidth]{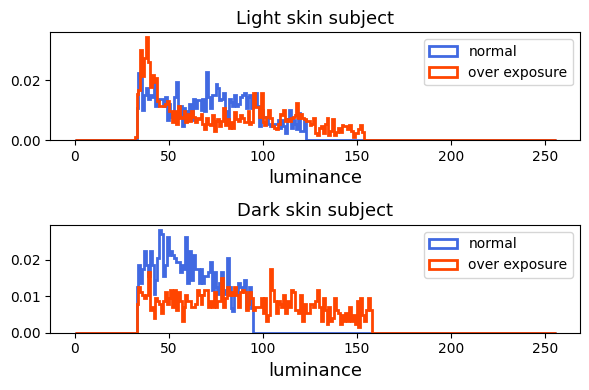}
         \vspace{-0.45cm}
         \caption{Sclera}
         \label{fig:ox-lumi-sclera}
     \end{subfigure}
     \vspace{-0.25cm}
    \caption{Normal vs. Over exposure  for face skin vs. sclera}
    \label{fig:ox-lumi}
\end{figure}

Figure \ref{fig:ox-lumi} shows the luminance histograms of normal vs. over exposed, and face skin vs. sclera images of two subjects. Figure \ref{fig:ox-lumi-face} on the left shows the histograms for the face skin and Figure \ref{fig:ox-lumi-sclera} on the right shows the ones for the sclera. It can be seen from the histograms that, as expected, over exposure increases the luminance values and pushes them toward the right, and this is true for both face skin and sclera.

Column 1 in Table \ref{tbl:distances} shows that the distance between the two sclera histograms is much smaller than the distance between the two face skin histograms of the over exposed images on two distance measures.

\begin{figure}[h!]
     \centering
     \begin{subfigure}[b]{0.45\linewidth}
         \centering
         \includegraphics[width=\linewidth]{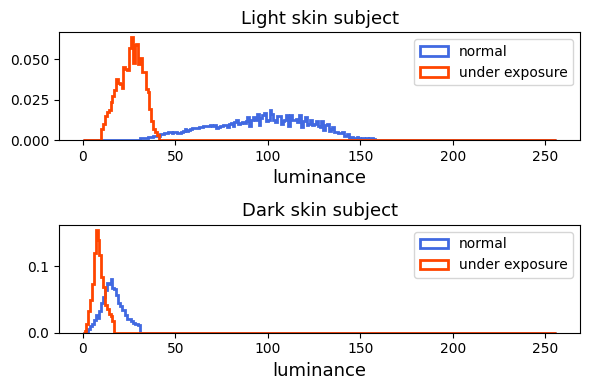}
         \vspace{-0.45cm}
         \caption{Face}
         \label{fig:ux-lumi-face}
     \end{subfigure}
     \begin{subfigure}[b]{0.45\linewidth}
         \centering
         \includegraphics[width=\linewidth]{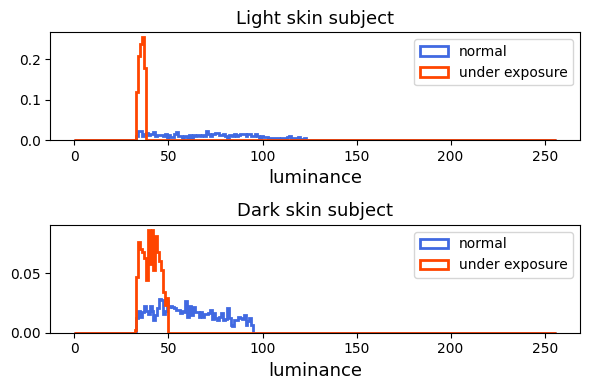}
         \vspace{-0.45cm}
         \caption{Sclera}
         \label{fig:ux-lumi-sclera}
     \end{subfigure}
     \vspace{-0.25cm}
    \caption{Normal vs. Under exposure for face skin vs. sclera}
    \label{fig:ux-lumi}
\end{figure}

Figure \ref{fig:ux-lumi} shows the luminance histograms of normal vs. under exposed, and face skin vs. sclera images of two subjects. Figure \ref{fig:ux-lumi-face} on the left shows the histograms for the face skin and Figure \ref{fig:ux-lumi-sclera} on the right shows the ones for the sclera. It can be seen from the histograms that under exposure decreases the luminance values and limits their range, and this is true for both face skin and sclera. Column 2 in Table \ref{tbl:distances} shows that the distance between the two sclera histograms is much smaller than the distance between the two face skin histograms of the under exposed images on two distance measures.

\section{Conclusion and Future Work}

Face image quality assessment is important for obtaining high face recognition performance. Face recognition systems, as well as FIQA algorithms, should be fair, and perform consistently across different demographic groups. This work investigated the use of the eye sclera as an alternative to face skin for assessing some quality components of a face image. It examined closely three face image quality components, namely, dynamic range, over- and under exposure, and implemented corresponding quality assessment algorithms. The algorithms are adapted from ISO/IEC CD3 29394-5 and have been adjusted to work with the face skin and sclera data. The evaluation results show that the eye sclera has very similar predictive capacity in assessing the quality of a face image with compared to the face skin. Thus, it can be used as a supplementary or an alternative method for a fully skin tone agnostic mechanism for assessing these quality factors to make sure that no bias is present in the assessment.

Future work can look at extending this approach to other quality components and examine whether the eye sclera has the same predictive capacity as the face skin with regard to other aspects of the face image.

\section*{Acknowledgement}
This work was supported by the European Union’s Horizon 2020 Research and Innovation Program under Grant 883356.

\end{document}